\documentclass[11pt,a4paper]{article}
\usepackage[hyperref]{acl2021}
\usepackage{times}
\usepackage{latexsym}
\usepackage{graphicx}
\usepackage{subfig}

\usepackage{booktabs}
\usepackage{multirow}

\usepackage{pifont}

\usepackage{amsmath}
\usepackage{breqn}
\usepackage{microtype}

\aclfinalcopy 

\graphicspath{ {./images/} }

\newcommand{\err}[1]{\textsubscript{~$\pm$#1}}

\title{Volta at SemEval-2021 Task 6: Towards Detecting Persuasive Texts and Images using Textual and Multimodal Ensemble}

\author{Kshitij Gupta\thanks{\enspace The authors have contributed equally.} \qquad Devansh Gautam\footnotemark[1] \qquad Radhika Mamidi \\
  International Institute of Information Technology Hyderabad \\
  \texttt{\{kshitij.gupta,devansh.gautam\}@research.iiit.ac.in},\\ \texttt{radhika.mamidi@iiit.ac.in}
}

\date{}

\begin{document}
\maketitle
\begin{abstract}
Memes are one of the most popular types of content used to spread information online. They can influence a large number of people through rhetorical and psychological techniques. The task, \textit{Detection of Persuasion Techniques in Texts and Images}, is to detect these persuasive techniques in memes. It consists of three subtasks: (A) Multi-label classification using textual content, (B) Multi-label classification and span identification using textual content, and (C) Multi-label classification using visual and textual content. 
In this paper, we propose a transfer learning approach to fine-tune BERT-based models in different modalities.
We also explore the effectiveness of ensembles of models trained in different modalities. We achieve an F1-score of 57.0, 48.2, and 52.1 in the corresponding subtasks.
\end{abstract}

\section{Introduction}
Memes are text superimposed on graphics that convey messages in the form of jokes, sarcasm, etc. In the current era of the internet and social media, they are very quick to spread. If used as a part of a disinformation campaign, it can be quite tricky to notice the agenda behind them and has the potential to influence a large mass of people without them realizing it~\citep{Muller2018,brazil,Glowacki:18}.

To this end, SemEval 2021 Task 6~\citep{SemEval2021-6-Dimitrov} focuses on identifying such persuasive techniques~\citep{Miller} in a multimodal (visual-linguistic) setting. It consists of three subtasks that enable the participants to study the problem in each modality. Only the English textual cues are used in tasks A and B, while the visual cues are also used in task C. Task B is a modification of task A which further requires predicting the spans for each identified technique as well.

\begin{figure}%
\centering
    \subfloat{
    \includegraphics[width=\columnwidth,height=100pt,keepaspectratio]{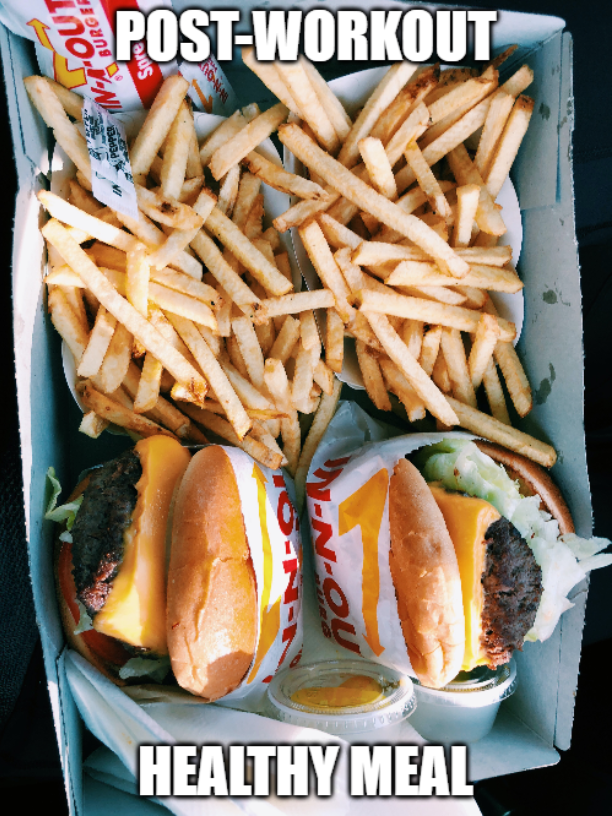}
    }%
    \\
    \subfloat{{
    \includegraphics[width=\columnwidth,height=80pt,keepaspectratio]{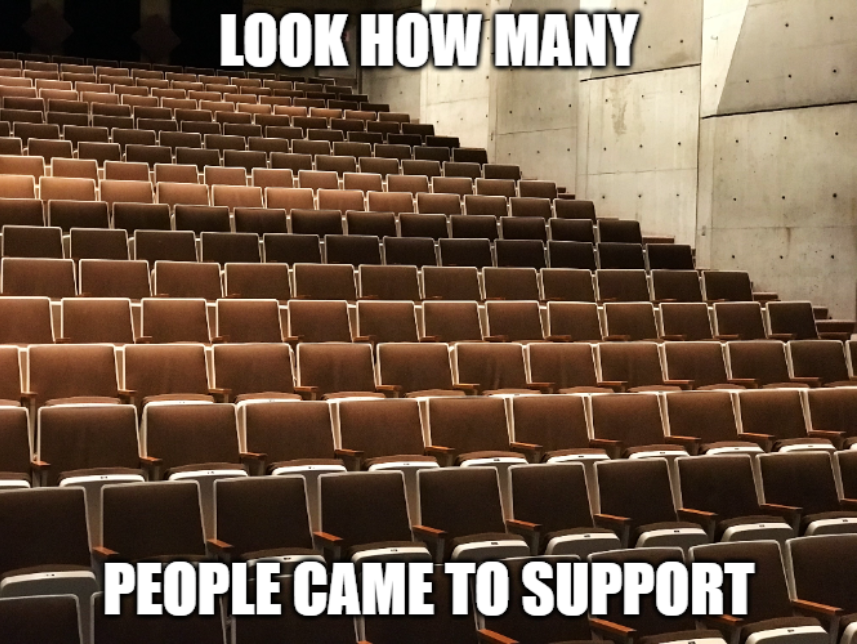}
    }}%
    \subfloat{{
    \includegraphics[width=\columnwidth,height=80pt,keepaspectratio]{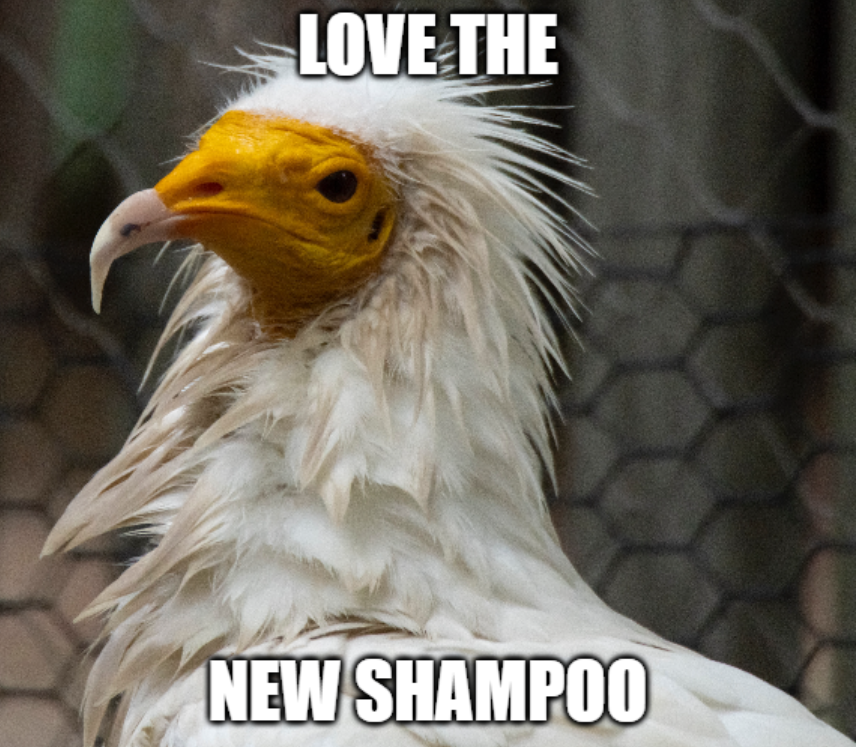}
    }}%
    \caption{Sample memes demonstrating the multimodal setting}%
    \label{fig:samples}%
\end{figure}

Meme classification is a multimodal problem that often requires visual and textual cues to convey a message. Memes can convey very different meanings if either of the cues is removed. A few samples are shown in figure \ref{fig:samples} to demonstrate the importance of visual and textual cues for classification.

We experiment with BERT~\citep{devlin-etal-2019-bert} based unimodal models for tasks A and B. Since they are state-of-the-art models for natural language understanding, they are a good choice for understanding the complex propaganda techniques in texts. 
Transformers~\citep{NIPS2017_3f5ee243} have limited sequence length, which limits their performance on longer data, but in the case of memes, the textual content is also very limited.

For task C, we experiment with Visual-Linguistic (VL) models like UNITER~\citep{chen2020uniter}, VisualBERT~\citep{li2019visualbert}, LXMERT~\citep{tan2019lxmert} for the cross-modal understanding of memes. We further explore the effectiveness of ensembling models trained in different modalities.

The code for all subtasks is available at \url{http://github.com/kshitij98/multimodal-propaganda.git}

\section{Background}
Propaganda aims to push biased agendas to influence people's mindsets. It is successful in achieving its goal by hiding its way through any of the numerous media platforms available in the current world. A major factor behind the success of such campaigns is the boom of the internet and social media in recent years. Another factor being the difficulty to spot such techniques manually because of the high volume of text produced and the unnoticeable nature of such content. With the recent interest of the research community in "fake news," the detection of persuasive techniques or highly biased texts has emerged as an active research area. Some of the previous work in this direction analyzes the general pattern of propaganda \citep{10.1145/3140565, 10.1145/2757401.2757408}, performs analysis at a document level \citep{rashkin-etal-2017-truth, Barron-CedenoMJ19} and a fine-grained analysis of the text \citep{da-san-martino-etal-2019-findings, da-san-martino-etal-2020-semeval}. However, most previous work analyses the techniques in a textual unimodal setting only. This work studies propaganda techniques in a new age domain like memes.

Meme classification task can be considered a combined VL multimodal problem. It is similar to the currently popular VL problems like Visual Question Answering \citep{VQA}, Visual Commonsense Reasoning \citep{Zellers_2019_CVPR} and Visual Entailment \citep{xie2019visual}, as we have to classify semantically correlated text with that of the visual content in the image. Hence a cross-modal approach under vision and text should perform better than unimodal architectures.
Basic VL approaches are based on simple fusion techniques in the form of early or late fusion to correlate unimodally trained visual and textual models. 
However, in an ideal scenario, a multimodally trained model should be more effective in detecting persuasive techniques in memes. With the rising interest in VL problems, recent work attempts to study similar problems in a VL multimodal setting \citep{Gomez_2020_WACV, kiela2020hateful, suryawanshi-etal-2020-multimodal}.

\paragraph{Data Description}

The dataset consists of 951 memes in total, which is further divided into train/ dev/ test splits. All the tasks have the same set of memes in their training sets, but the labeling differs for each of them.
For task A, only the textual cues were used to identify the techniques. For task B, the spans of each technique were further detected. For task C, more techniques were identified using visual cues.

\begin{figure}
    \centering
    \includegraphics[width=\columnwidth,height=220pt,keepaspectratio]{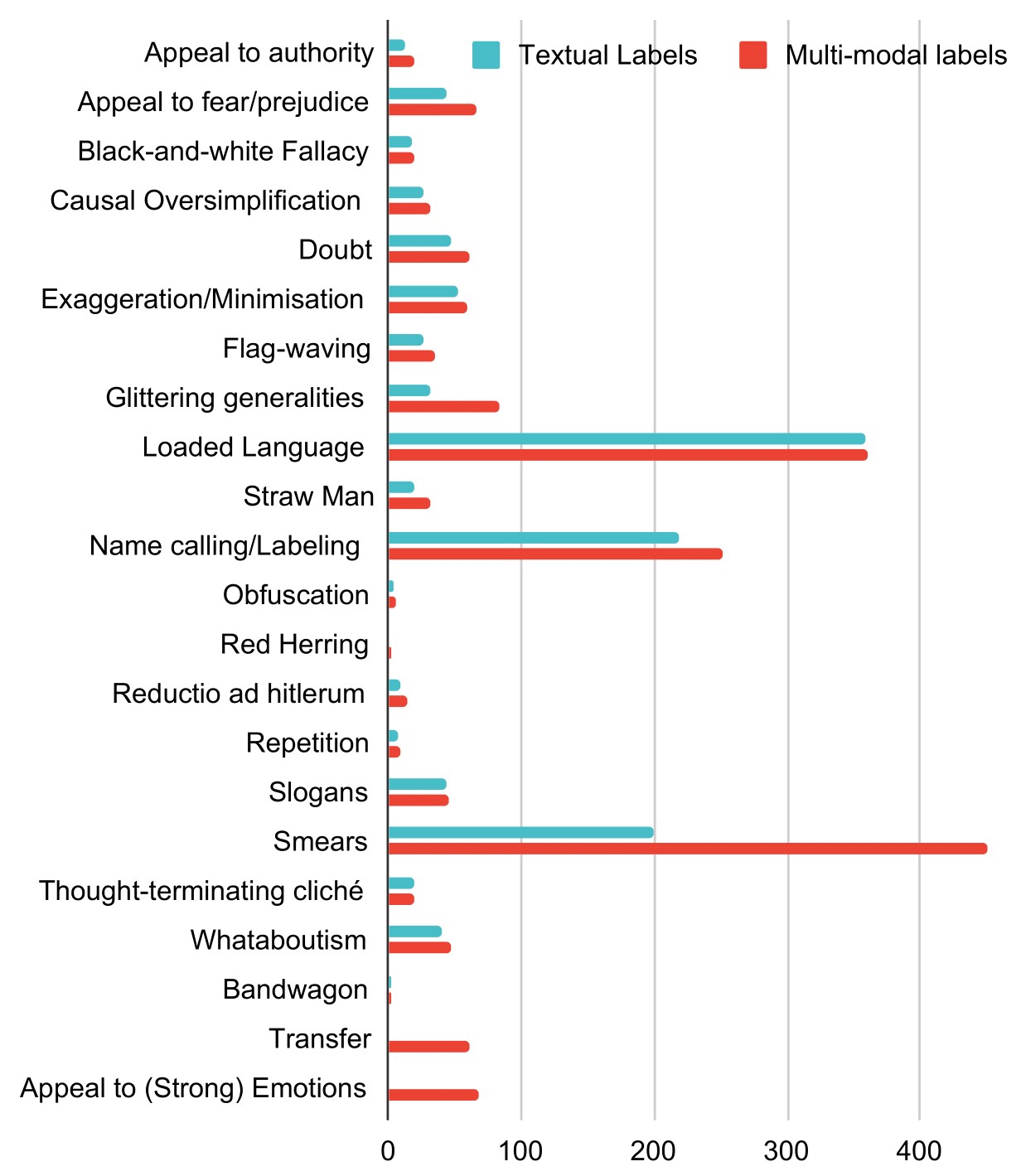}
    \caption{Data Distribution of the labels in the training set}
    \label{fig:data}
\end{figure}

The distribution of the labels is illustrated in figure \ref{fig:data}. Detailed information of the dataset can be found in the task description paper \citep{SemEval2021-6-Dimitrov}.




\section{System Overview}

Systems proposed for all the tasks use BERT-based models with task-specific modifications.

\begin{figure*}
\centering
    \subfloat[\centering Task A\label{fig:modela}]{{\includegraphics[height=5cm]{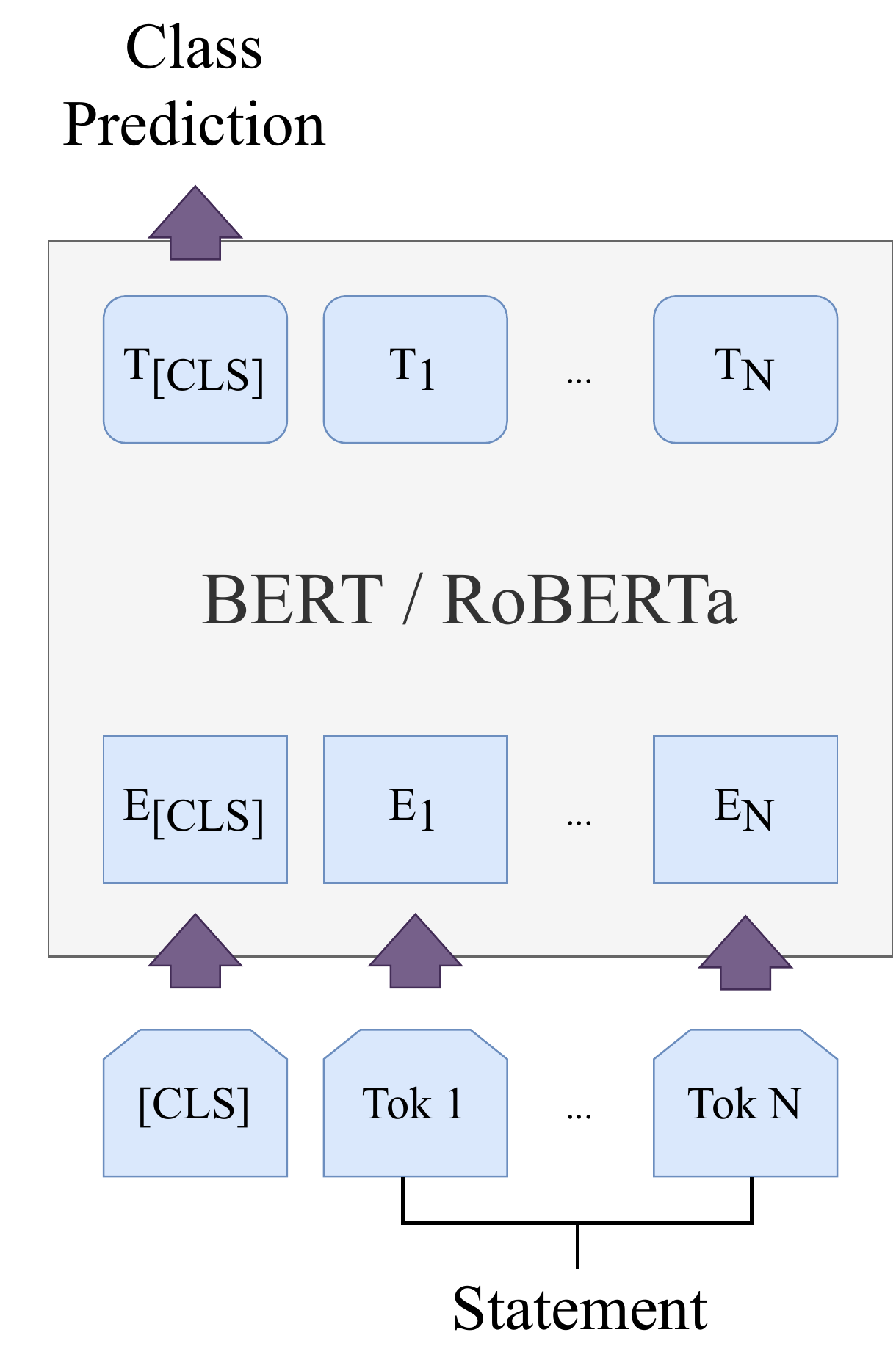} }}
    \quad
    \subfloat[\centering Task B\label{fig:modelb}]{{\includegraphics[height=5cm]{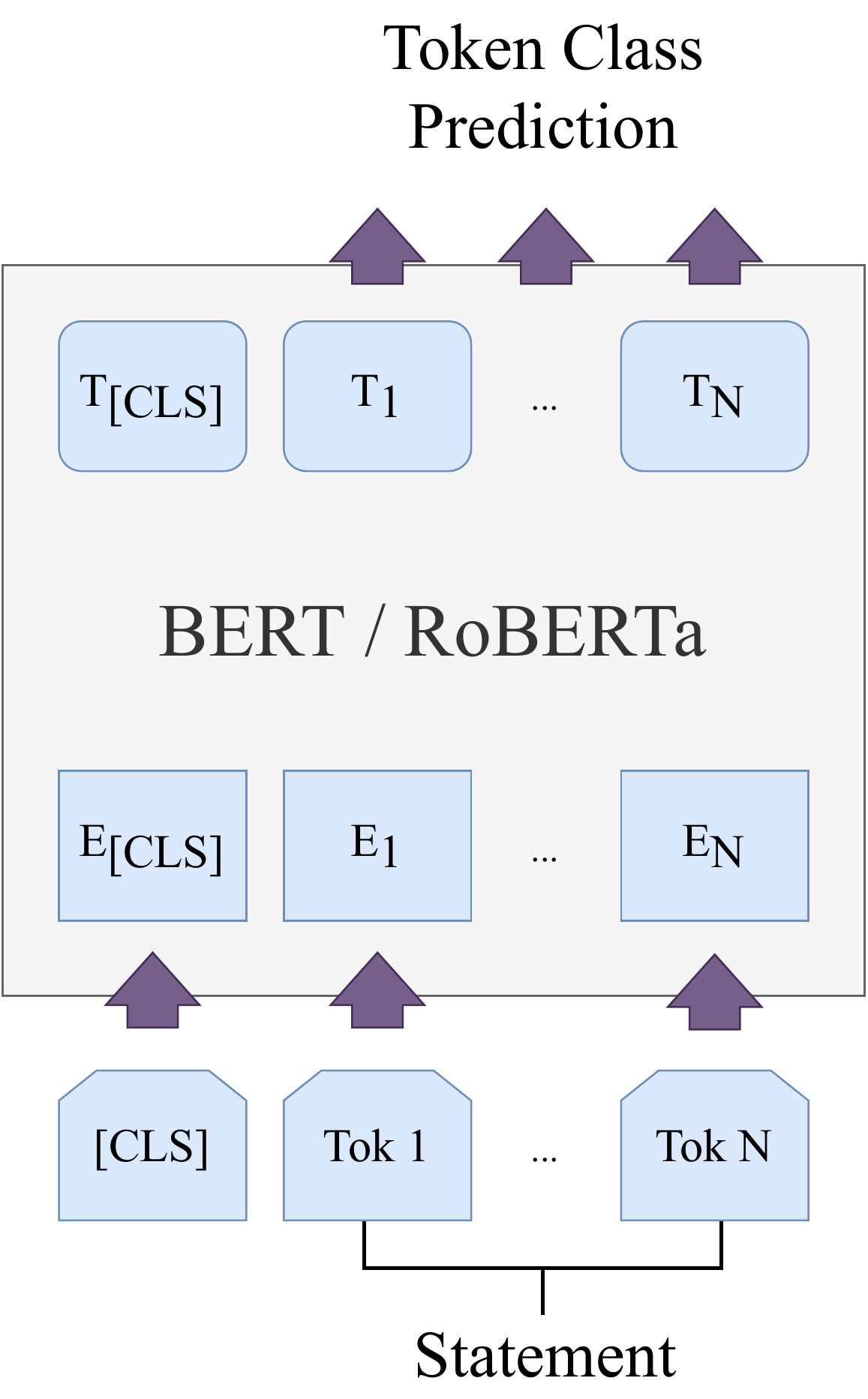} }}
    \quad
    \subfloat[\centering Task C\label{fig:modelc}]{{\includegraphics[height=5.6cm]{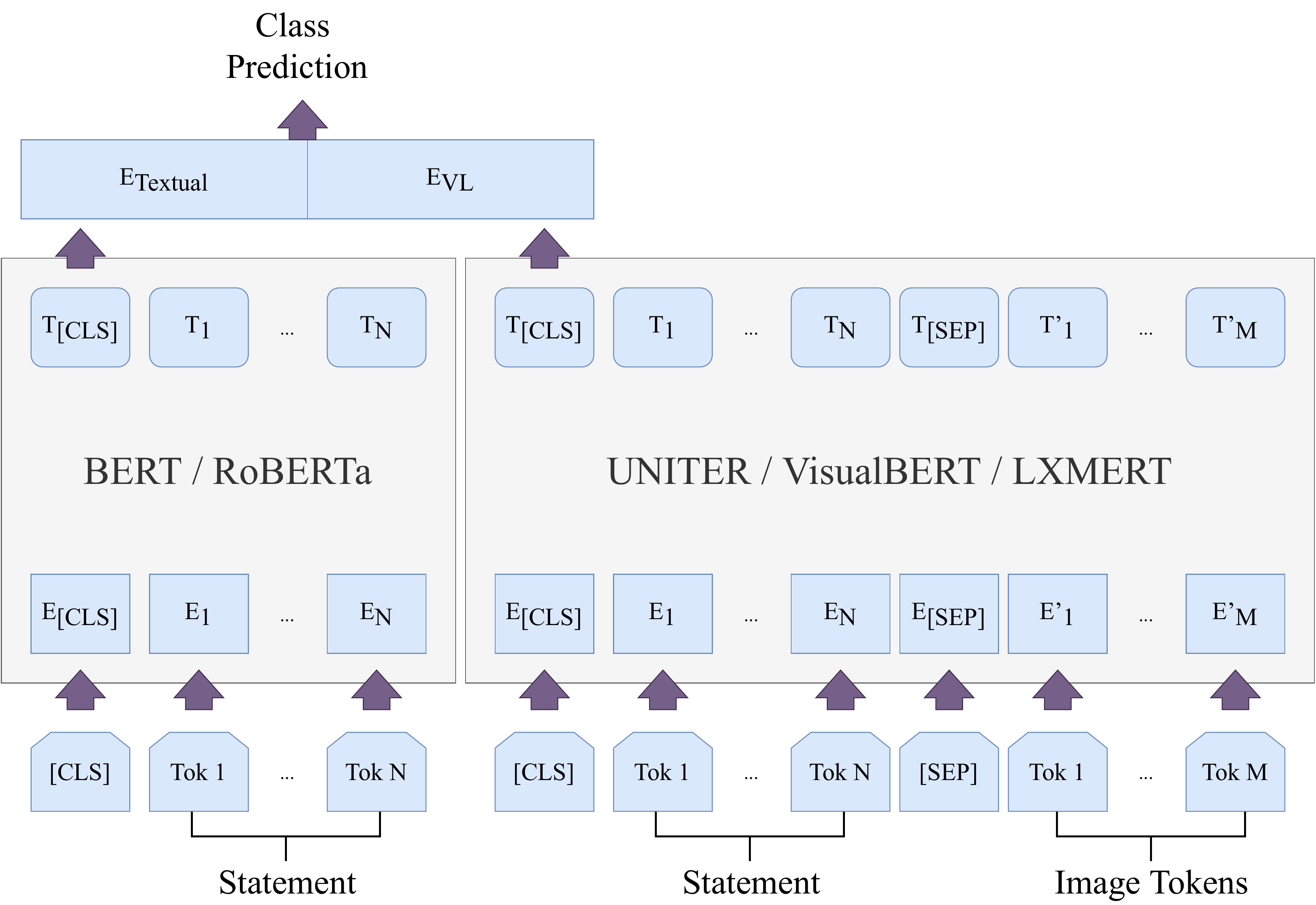} }}

    \caption{Proposed architectures for the given subtasks}
    \label{fig:model}%
\end{figure*}

The proposed systems are explained in detail below:

\subsection{Task A - Text Classification}
We use transfer learning to fine-tune BERT-based models for this task. To fine-tune the models, we experiment with {\small BASE} and {\small LARGE} variants of BERT and RoBERTa. We attach a feed-forward network on the \texttt{[CLS]} token embedding with two linear layers having the model's default dropout and $Tanh$ activation layer in between.
The model architecture is illustrated in figure \ref{fig:modela}. We apply a standard Binary Cross Entropy loss to train this model with the hyperparameters mentioned in table \ref{tab:hyperparameters}. Since there is a substantial class imbalance in the dataset, we add weights to the positive samples in the loss function by using the following equation: 

\resizebox{0.75\columnwidth}{!}{\begin{minipage}{\columnwidth}
\begin{align} \label{eqn:loss} \begin{aligned} \ell(\mathbf{x}, \mathbf{y}) & = -\frac{1}{Nd}\sum_{n=1}^N \sum_{k=1}^d\big[p^k y_n^k \log x_n^k + (1 - y_n^k)\log(1 - x_n^k) \big] \\ p^k & = \frac{1}{f^k}(|K| - f^k)\\ \end{aligned} \end{align}
\end{minipage}
}

Where $N$ is the batch size, $n$ index denotes $n^{th}$ batch element, $d$ is the number of classes, $f$ stands for a vector of class absolute frequencies calculated on the train set, $\mathbf{x}$ is the output vector from the last Sigmoid layer, $\mathbf{y}$ is a vector of multi-hot encoded ground truth labels and $|K|$ is the size of the train set.

We finally use the standard Sigmoid activation function to compute probabilities for each label.

\subsection{Task B - Span Identification and Text Classification}

We experiment with the same BERT-based encoders for this task.

\paragraph{Pre Processing}
Since the spans are given on character-level in the dataset and the transformer models run on token-level, we transform all the spans to token-level by taking the intersection with the tokenized input. Further, we train the model with the obtained token-level labels as targets.

\paragraph{Model}
We model the problem statement as a token-level \textit{multi-label} classification problem for span identification. The model architecture is illustrated in figure \ref{fig:modelb}. We use the same classifier to classify all token embeddings.
To handle class imbalance, we apply weighted Binary Cross-Entropy loss with class weights as mentioned in equation \ref{eqn:loss}

\paragraph{Post Processing}
Since all tokens of a word should belong to the same set of classes, we merge all the tokens of each word and assign a union of those classes to the chosen word.

Finally, all the words are classified using the following equation where $W_{i}$ is the set of labels of the $i^{th}$ word, $L_{i, j}$ is the set of labels of the $j^{th}$ token of the $i^{th}$ word and ${N_{i}}$ is the number of tokens of the $i^{th}$ word: 

    $$ W_{i} = \bigcup\limits_{j=1}^{N_{i}} L_{i, j} $$

We do not apply loss on special tokens so that the classifier is not misled while training.

\subsection{Task C - Visual-Linguistic Classification}
We experiment with Visual-Linguistic (VL) models for this task. Since meme classification is a multimodal task, a multimodal transformer architecture should be a good choice. 

\paragraph{Pre Processing}

The image tokens used in the VL models are extracted using Faster R-CNN \citep{NIPS2015_14bfa6bb} and a fixed number of image tokens are created by extracting the features of the top 36 regions of interest.

\begin{table}[t]
\centering
\resizebox{\columnwidth}{!}{%
\begin{tabular}{@{}lcccc@{}}
\toprule
\multicolumn{1}{c}{\multirow{2}{*}{\textbf{Parameter}}} & \multirow{2}{*}{\textbf{Task A}} & \multirow{2}{*}{\textbf{Task B}} & \multicolumn{2}{c}{\textbf{Task C}} \\ \cmidrule(l){4-5} 
\multicolumn{1}{c}{} &  &  & \textbf{VL} & \multicolumn{1}{c}{\textbf{Ensemble}} \\ \midrule
\textbf{Dropout} & 0.1 & - & - & - \\
\textbf{Max Sequence Length} & 128 & 512 & 128 & 128 \\
\textbf{Batch Size} & 8 & 8 & 16 & 32 \\
\textbf{warmup} & - & - & 0.1 & 0.1 \\
\textbf{Learning Rate} & 1e-05 & 1e-05 & 1e-05 & 1e-05 \\
\textbf{patience} & 50 & 50 & 50 & 50 \\
\textbf{Weight Decay} & - & 0.01 & 0.01 & 0.01 \\
\textbf{Optimizer} & Adam & AdamW & BertAdam & BertAdam \\

\bottomrule
\end{tabular}%
}
\caption{Hyperparameters}
\label{tab:hyperparameters}
\end{table}

\paragraph{Model}
We first experiment with training the VL models directly. Although the VL models should perform better in the presence of extra cues for classification, poor performance is observed compared to the powerful textual-only models. 

On further analyzing the problem, we observe that propaganda detection is a complex semantic problem, and often the classes can be detected by using just the text. So, we experiment with the textual models and ignore the visual cues in the data. Surprisingly, the textual model outperforms the VL model when trained to learn multimodal labels using just the textual cues. We further study both models to better understand the learnings of each of them (see Section \ref{results}) and continue experimentation with ensembling both models.

We propose an ensemble of multimodal transformers like UNITER, VisualBERT, LXMERT with unimodal transformers like BERT and RoBERTa to help the classification model as each architecture can focus on their domains and later merge those embeddings. Rather than using a naive average ensembling method, we propose our own model, illustrated in figure \ref{fig:modelc}. Our model concatenates the \texttt{[CLS]} token embeddings from transformers in each modality and then applies classification on top of the concatenated vector. The base encoders are fine-tuned unimodally and then frozen while training the ensemble classifier. To train the textual model unimodally, we train the textual model with textual labels only to learn relevant features.

\section{Experimental Setup}

\subsection{Implementation}

\begin{table*}
\centering
\begin{tabular}{lcccccc}

\toprule
\textbf{Model} & \multicolumn{2}{c}{\textbf{Task A}} & \multicolumn{2}{c}{\textbf{Task B}} & \multicolumn{2}{c}{\textbf{Task C}}\\
& Dev Set & Test Set & Dev Set & Test Set & Dev Set & Test Set \\

\midrule
BERT~\textsubscript{BASE} & $64 \err{0.7}$ & $52.2 \err{2.5}$ & $53.1 \err{1}$ & $45.6 \err{1.5}$ & $63.3 \err{1.2}$ & $51.9 \err{1}$ \\
BERT~\textsubscript{LARGE} & $62.8 \err{0.8}$ & $\textbf{54.8} \err{1.5}$ & $53.5 \err{2.1}$ & $44.7 \err{2.5}$ & $62.7 \err{1.4}$ & $52.5 \err{0.9}$ \\
RoBERTa~\textsubscript{BASE} & $61 \err{0.8}$ & $51.2 \err{1.2}$ & $53.2 \err{0.6}$ & $43.9 \err{0.9}$ & $61.5 \err{1.1}$ & $49.8 \err{1.2}$ \\
RoBERTa~\textsubscript{LARGE} & $\textbf{64.7} \err{1.1}$ & $53.2 \err{3.9}$ & $\textbf{58.5} \err{2.1}$ & $\textbf{47.6} \err{1.5}$ & $63.9 \err{1.1}$ & $54.2 \err{1}$ \\

\midrule
UNITER &  &  &  &  & $64.9 \err{0.2}$ & $49.2 \err{0.6}$ \\
VisualBERT &  &  &  &  & $57.8 \err{0.6}$ & $45.8 \err{0.7}$ \\
LXMERT &  &  &  &  & $54.5 \err{0.3}$ & $44.4 \err{0.0}$ \\
\midrule

BERT~\textsubscript{BASE} + UNITER &  &  &  &  & $66.1 \err{0.2}$ & $52.9 \err{0.8}$ \\
RoBERTa~\textsubscript{LARGE} + UNITER &  &  &  &  & $\textbf{67.3} \err{0.9}$ & $\textbf{54.9} \err{0.6}$ \\
BERT~\textsubscript{BASE} + VisualBERT &  &  &  &  & $63.1 \err{0.3}$ & $53.1 \err{0.5}$ \\
RoBERTa~\textsubscript{LARGE} + VisualBERT &  &  &  &  & $64.4 \err{0.5}$ & $52.7 \err{1.9}$ \\

\bottomrule

\end{tabular}
\caption{\label{performance}
Mean and std of F1-micro scores computed from 10 runs of the mentioned models. Test set values are reported after choosing the best model for the dev set using early stopping.
}
\end{table*}

We use HuggingFace \footnote{\hspace{1pt}\includegraphics[height=6.5pt]{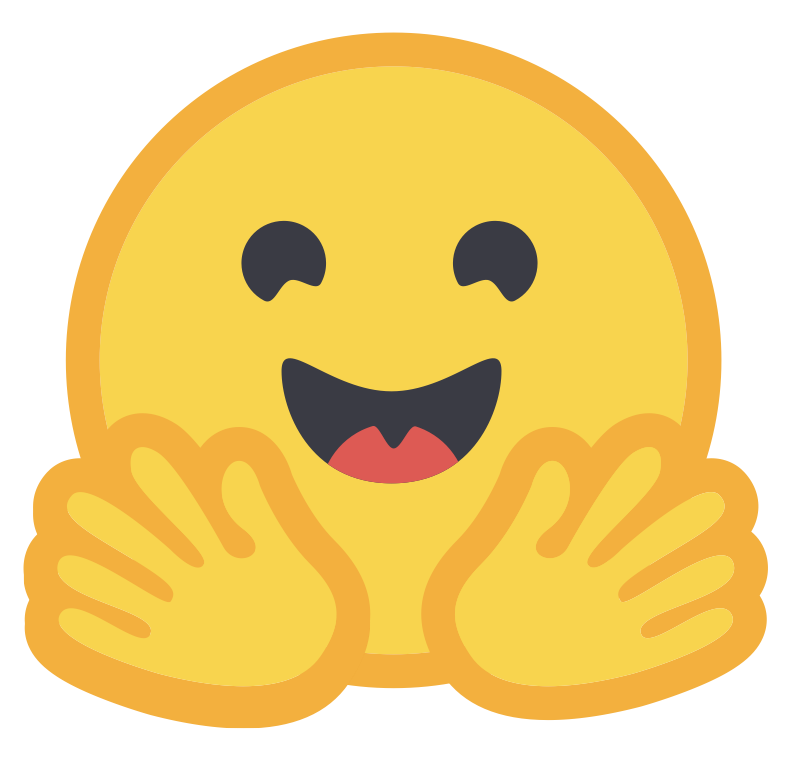} Transformers, v4.2.0, \url{https://huggingface.co/transformers/}} library \citep{wolf-etal-2020-transformers} to experiment with textual models, and use the official implementations of the multimodal transformers in PyTorch\footnote{PyTorch, v1.7.1, \url{https://pytorch.org/}}~ \citep{NEURIPS2019_9015}. The models were trained with the default hyperparameters with an exception for the parameters mentioned in table \ref{tab:hyperparameters}. All experiments were conducted using Nvidia GeForce RTX 2080 Ti GPU.

For training the models, we use the same train/ dev/ test split as mentioned in the task, which is in the ratio 10:1:3 with 951 samples in total. We use the best-performing model by comparing the F1-micro scores on the validation set for all our experiments.

\subsection{Evaluation Metrics}

For tasks A and C, the F1-micro score is used as the main performance metric. Due to various low resource classes in the dataset, F1-macro was also used to give weight to the smaller classes in the performance metric, but it was revealed to be highly impacted by small variations in the model.

For task B, we use the official evaluation metric as defined in the task. 
As the task is a multi-label sequence tagging task, the standard micro-averaged F1 is modified to account for partial matching between the spans. 

\section{Results}
\label{results}
We conduct several experiments to compare the performance of models on each of the tasks. Detailed information can be seen in table \ref{performance}.

For task A, RoBERTa~\textsubscript{LARGE} is the best performing model on the task. Although the average performance of the model is worse than BERT~\textsubscript{LARGE} on the test set, the maximum performance is still better.

Similarly, for task B, RoBERTa~\textsubscript{LARGE} outperforms other models by a large margin. 

For task C, we trained the textual models on visual-linguistic labels for a fair comparison with the other VL models. Unexpectedly, they also show decent performance and surprisingly perform better than the multimodal transformers. Finally, the ensemble models are trained with their textual model trained on textual labels and the VL model trained on VL labels. They outperform their unimodal components and end up with a fair performance increase in each of the cases.

During the post-evaluation phase, we conduct more experiments and report the corresponding performances on the test set for the best-validated models on the dev set. We realize that our submitted model performed worse on the test set in task C because the checkpoint used for test set predictions was different from the best validation checkpoint.

The final values we achieve on the test set by using the best validation set checkpoints are 57.56, 48.23, and 55.68 by using RoBERTa~\textsubscript{LARGE}, RoBERTa~\textsubscript{LARGE} and RoBERTa~\textsubscript{LARGE} + UNITER, respectively on all the subtasks.

We observe that all BERT-based models give a decent performance for all tasks. An interesting insight we get from the visual-linguistic task is that the problem is not completely multimodal. Several of the persuasive techniques can be identified by using just the textual cues, which is evident from figure \ref{fig:data} as well. While textual models like RoBERTa are pretrained on a much larger textual dataset and are able to learn more complex semantics from the data, VL models suffer when used in textual-only domains. We carry out experiments to further study the differences in these models with different types of labels and propose an architecture that benefits from both models.

\begin{figure}
    \centering
    \includegraphics[width=\columnwidth,height=\textheight,keepaspectratio]{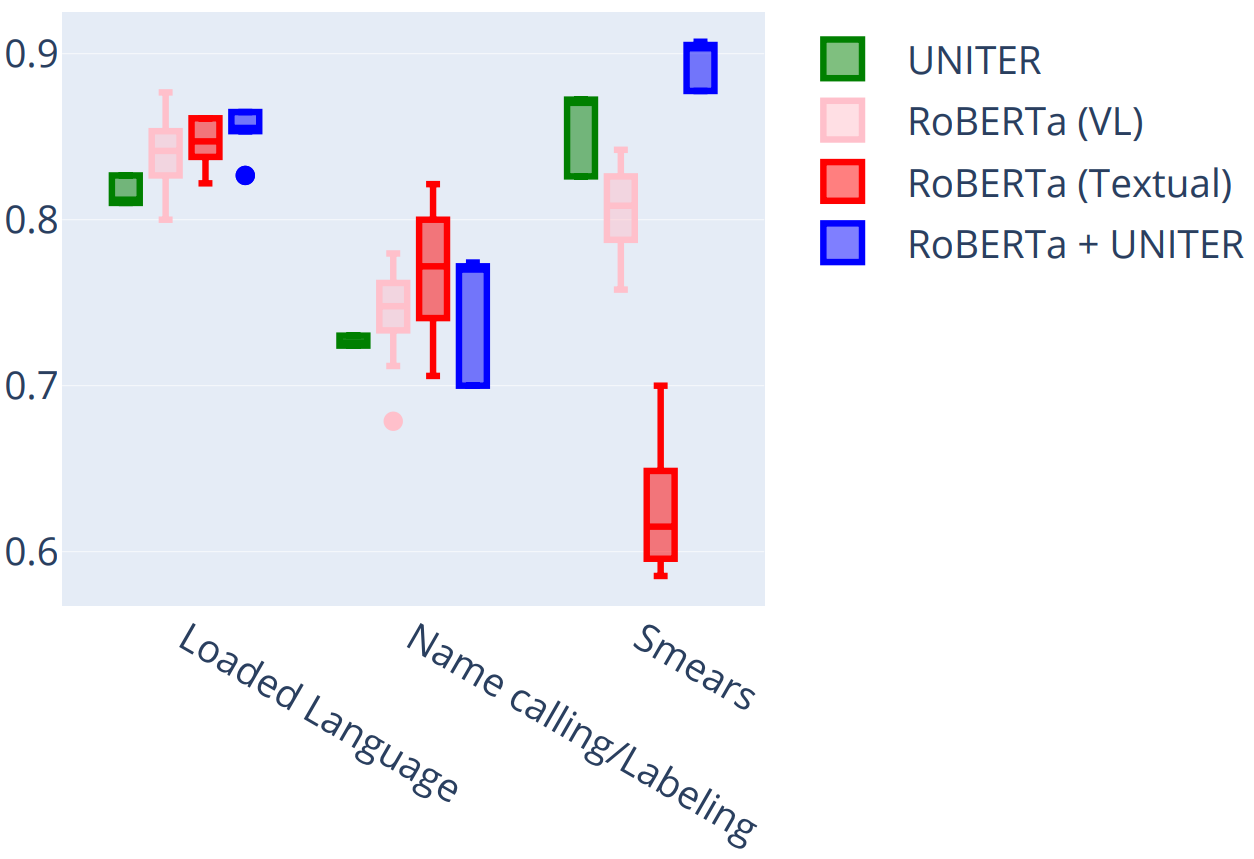}

    \caption{Comparison of different models in the 3 most occurring classes. F1-micro scores of 10 runs are computed on the dev set.}
    \label{fig:comparison}
\end{figure}

A class-wise comparison is carried out for all models. Since several of the classes have very low positive samples, it is difficult to draw conclusions because of the high variance in performance due to different model initializations, so we compare the most frequently occurring labels only. The difference in the performance of different models is illustrated in figure \ref{fig:comparison}. For this comparison, we train the textual model on multimodal and textual labels to compare the capability of the models. Ideally, the textual models should be trained on textual labels only to learn more relevant features. The ensemble is trained with a textual model trained on textual labels and a VL model trained on VL labels. The comparison shows that the multimodal models are performing much better for detecting \textit{Smears}, which has various samples which require visual cues as well. Another observation is that RoBERTa is performing better on \textit{Loaded Language} and \textit{Name calling / Labelling}. Training an ensemble of the textual model and the multimodal is helping the model perform better in all classes.

\begin{figure}
    \centering
    \includegraphics[width=\columnwidth,height=\textheight,keepaspectratio]{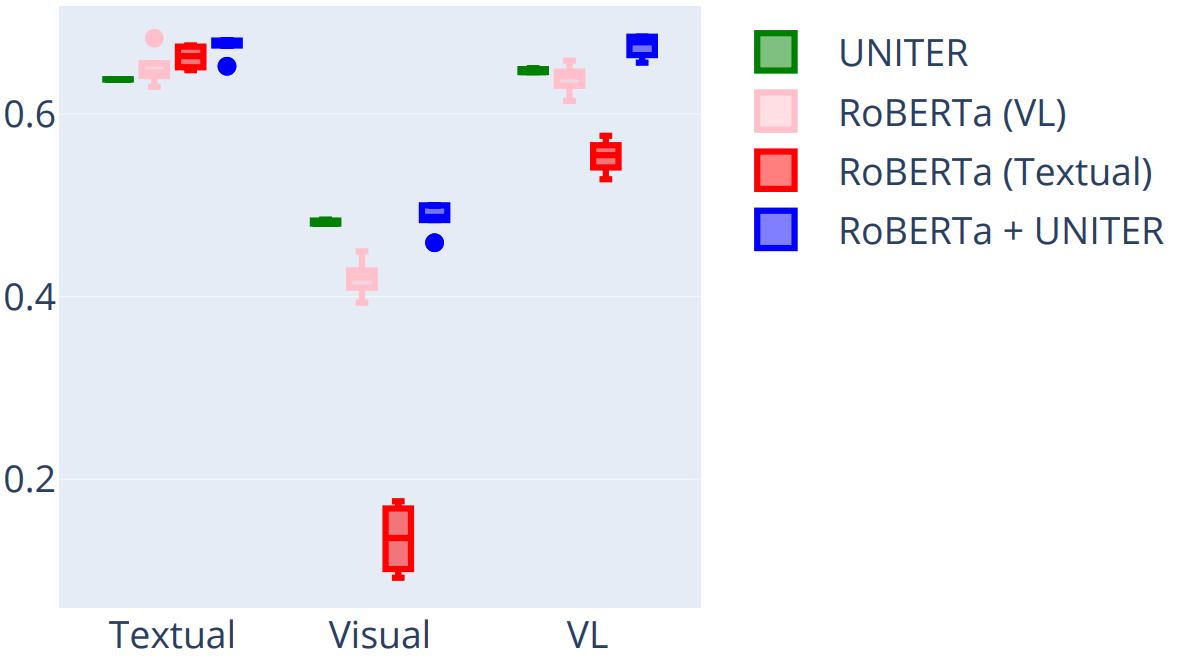}

    \caption{Comparison of different models in different modes. F1-micro scores of 10 runs are computed on the dev set.}
    \label{fig:modal-performance}
\end{figure}

To further study the models, we report the performance for each of them in different modalities. To measure the textual performance of the models, we compare the sets of labels in task A and C to shortlist labels which were identified only after the presence of visual cues. We do not consider those predictions and calculate the F1-micro score for the remaining subset of the predictions. Similarly, for measuring the visual performance, we do not consider the labels which were identified by just using the textual cues. The performance of several runs of the models is compared in figure \ref{fig:modal-performance}. The comparison also supports the claim that ensembling models trained in different modalities help to learn from the best of both worlds.

\section{Conclusion}

Although detecting persuasive techniques in memes is a multimodal problem, often, most of the techniques can be identified by just using the textual cues from the meme. Since VL models are still in their nascent stages, powerful textual models help with solving the problem at hand.
Future work can be done to improve these kinds of problems that are not multimodal in the truest sense. The ensembling method used in our model is very basic; better architectures can be explored to continue this line of work.

\section*{Acknowledgments}

We thank the organisers of the shared task for their effort, and the anonymous reviewers for their comments.

\bibliographystyle{acl_natbib}
\bibliography{anthology,acl2021}

\end{document}